\documentclass[runningheads]{llncs}

\usepackage{eccv}

\usepackage{caption}
\newsavebox{\mytablebox}
\usepackage{eccvabbrv}
\usepackage{makecell}
\usepackage{multirow}
\usepackage{verbatim}
\usepackage{graphicx}
\usepackage{booktabs}

\usepackage[accsupp]{axessibility}  

\usepackage{hyperref}

\usepackage{orcidlink}

\begin{document}
\title{RBE-Flow: Recurrent Bayesian Estimation on Feature Manifolds for Cross-Modal Registration} 
\titlerunning{RBE-Flow for Cross-Modal Registration}
\author{Mengzhu Ding\inst{1}\orcidlink{0009-0008-5158-4823} \and
Xin Song\inst{1}\orcidlink{0009-0006-3379-4099} \and
Xiaoke Ding\inst{1}\orcidlink{0009-0006-4109-1983}
\and
Hongwei Ding\inst{1}\orcidlink{0000-0002-0851-1994}
\and \\
Xuecong Liu\inst{1}\thanks{Corresponding author.}\orcidlink{0000-0002-7911-6318}}
\authorrunning{M.~Ding et al.}
\institute{Northeastern University, China \\
\email{dingmengzhu@mails.neu.edu.cn, liuxuecong@qhd.neu.edu.cn}}
\maketitle

\begin{abstract}

Cross-modal image registration is essential for multi-sensor perception but remains fundamentally challenging due to severe non-linear radiometric discrepancies and geometric distortions. 
Existing deterministic matching methods lack uncertainty awareness, struggling to navigate the resulting highly non-convex optimization landscape and frequently accumulating errors in ambiguous regions.
In this paper, we propose RBE-Flow, a novel framework that reformulates dense cross-modal flow estimation as a closed-loop recurrent Bayesian estimation problem on learned feature manifolds. 
Diverging from standard feed-forward regression, RBE-Flow establishes a robust self-correcting mechanism by deeply coupling feature-metric non-linear optimization with probabilistic state updates. Specifically, a Recurrent Manifold Optimization (RMO) block iteratively generates flow observations and their associated uncertainties, which are then optimally assimilated into the prior state via an Uncertainty-Adaptive Probabilistic Update (UAPU) using deterministic sigma-point projection. Crucially, the resulting calibrated posterior covariance is fed back to adaptively regularize the damping of subsequent optimization steps, allowing the system to modulate its convergence based on predictive confidence. To ensure stable probabilistic training, we introduce a hybrid supervision scheme featuring a geometry-aware rectified NLL loss that structurally prevents variance collapse. Extensive experiments on challenging OSdataset, WHU-OPT-SAR, and RoadScene benchmarks demonstrate that RBE-Flow consistently achieves state-of-the-art performance, outperforming existing methods by a significant margin, particularly under strict sub-pixel criteria. Project page: \url{https://github.com/NEU-Liuxuecong/RBE-Flow}

\keywords{Cross-modal image registration \and Recurrent Bayesian Estimation \and Uncertainty Estimation \and Dense Correspondence \and Manifold Optimization}
\end{abstract}

\section{Introduction}
\label{sec:intro}

Cross-modal image registration aims to establish reliable geometric correspondences between images acquired by heterogeneous sensors (e.g., optical-SAR, visible-infrared)~\cite{jiang2021review, chen2025survey, li2025object, yang2023towards}. This capability supports a wide range of downstream applications, including remote sensing analytics~\cite{liu2022fast, sun2025gdros, li2024sardet, qin2025must}, 3D scene understanding~\cite{leroy2024grounding, edstedt2025colabsfm}, and autonomous navigation~\cite{gehrig2024low}, where complementary modalities provide robustness under challenging illumination or weather conditions. Compared to alignment within the same modality, cross-modal registration is fundamentally more challenging because the underlying sensing mechanisms induce serve non-linear radiometric discrepancies (e.g., intensity reversals, speckle noise, and spectral gaps)~\cite{ye2019cfog, xiao2024adrnet}. These discrepancies violate the assumptions behind standard photometric alignment and often weaken local visual evidence.

In particular, for optical-SAR pairs, intensity reversals and speckle noise often lead to inconsistent feature statistics, while for visible-infrared pairs, the spectral gap further reduces cross-modal feature consistency~\cite{mao2025cross, wu2024single}. Beyond appearance gaps, cross-modal registration frequently involves substantial geometric variations caused by viewpoint changes, scale differences, or local deformations~\cite{schusterbauer2025diff2flow, zhang2025adapting, li2025implicit}. These spatial deformations complicate pixel-level alignment and render traditional feature-based or intensity-driven signals unreliable~\cite{dai2024dsap, wu2025dfm}. These coupled factors give rise to a rugged, highly non-convex matching landscape riddled with local minima. Existing deterministic regression or standard optimization methods struggle to maintain stable refinement, succumbing to overconfident erroneous updates when faced with ambiguous cues~\cite{zeng2025uncertainty,hu2022uncertainty}.

To address these challenges, we propose a novel multi-modal registration network, RBE-Flow, that formulates cross-modal flow estimation as a closed-loop recurrent Bayesian estimation on feature manifolds. 
RBE-Flow starts from a coarse global initialization and performs refinement by minimizing a feature-metric non-linear least-squares energy, producing an iterative flow observation with associated uncertainty. We then apply an uncertainty-adaptive Bayesian posterior update via deterministic sigma-point projection, which fuses the new likelihood evidence with the prior belief and feeds back calibrated the posterior covariance to regulate subsequent optimization.

Our contributions are summarized as follows:

- We propose RBE-Flow, a unified closed-loop recurrent Bayesian estimation framework that integrates mathematical minimization with empirically derived uncertainty ensuring the robustness of the overall framework..

- We introduce the RMO block to generate local likelihood evidence by solving a feature-metric non-linear least-squares problem on the manifold, with uncertainty-aware damping to stabilize updates under cross-modal noise.

- We devise UAPU, an uncertainty-adaptive Bayesian posterior update that fuses the prior belief with RMO's observation and uncertainty via deterministic sigma-point projection, yielding a Minimum Mean Square Error (MMSE)-optimal fusion gain and posterior covariance to down-weight unreliable evidence in ambiguous, occluded regions.

- We design a hybrid supervision strategy with an $L_1$ loss for coarse initialization and a geometry-aware rectified NLL for recurrent refinement, where a dynamic variance lower bound prevents variance collapse and enforces uncertainty calibration against geometric error.

\section{Related Work}

\subsection{Feature Matching and Cross-Modal Registration}
Cross-modal registration entails establishing robust correspondences under severe radiometric differences, evolving through three paradigms:

\textbf{Hand-Crafted Matching.} Classical methods employ modality-invariant descriptors (e.g., RIFT~\cite{li2019rift}, OS-SIFT~\cite{xiang2018ossift}, CFOG~\cite{ye2019cfog}) using phase congruency or oriented gradients to handle intensity variations. Recent advanced geometric and intensity-invariant designs include POS-GIFT~\cite{hou2024pos}, MSG~\cite{zheng2025msg}, 
MMAR~\cite{liu2026multi},
MIRD~\cite{xuecong2024robust},
and shape-adaptive MIRD~\cite{liu2024shape}. However, these fixed descriptors inherently lack distinctiveness in low-texture regions and struggle with complex local deformations.

\textbf{Sparse Learnable Matching. } Deep learning shifted the focus toward data-driven sparse matching to overcome fixed descriptor limitations. SuperPoint~\cite{detone2018superpoint} and SuperGlue~\cite{sarlin2020superglue} pioneered graph neural networks for context-aware point matching. Advances, such as LightGlue~\cite{lindenberger2023lightglue}, OmniGlue~\cite{jiang2024omniglue}, and XFeat~\cite{potje2024xfeat}, have further enhanced efficiency and generalization. However, their reliance on repeatable keypoint detection frequently becomes a critical bottleneck across modalities with massive appearance gaps.

\textbf{(Semi-)Dense Learnable Matching. }To solve the detection bottleneck, detector-free methods infer correspondences directly from feature maps. LoFTR~\cite{sun2021loftr} and its variants~\cite{chen2022aspanformer, giang2023topicfm} use Transformers to capture long-range dependencies for semi-dense matching. Recently, models like XoFTR~\cite{tuzcuouglu2024xoftr} and RoMa~\cite{edstedt2024roma} have further adapted this paradigm for robust cross-modal and dense feature matching. Although powerful, extending them to dense cross-modal flow remains challenging due to the highly non-convex optimization landscape.

\subsection{Dense Flow Estimation and Geometric Refinement}
Dense flow estimation has become a dominant registration paradigm. Pioneered by RAFT~\cite{teed2020raft}, architectures like GMFlow~\cite{xu2022gmflow}, GMA~\cite{jiang2021learning}, FlowFormer~\cite{huang2022flowformer}, and DPFlow~\cite{Morimitsu_2025_CVPR} utilize recurrent correlation updates. To tackle cross-modal geometric transformations, recent frameworks like GAFF~\cite{liu2026gaff} and GDROS~\cite{sun2025gdros} introduce geometry-guided correlations while CRFT~\cite{liu2026crft} employs a consistent-recurrent Transformer for iterative spatial calibration. However, their implicit geometric constraints and learned updates limit robustness under large appearance gaps.

While GLU-Net~\cite{truong2020glu} handles large displacements via global-local correlations, its refinement implicitly approximates optimization. Conversely, embedding explicit differentiable least-squares solvers into deep networks (e.g., OptNet~\cite{amos2017optnet}, DROID-SLAM~\cite{teed2021droid}) proves highly effective. Motivated by the underexplored application of explicit second-order solvers to dense cross-modal flow, we incorporate a differentiable Levenberg-Marquardt (LM)-style solver for robust refinement.

\subsection{Uncertainty-Aware Refinement}
Estimating uncertainty is crucial for reliable registration~\cite{chen2025survey}. Existing approaches, such as PDC-Net~\cite{truong2021learning} and probabilistic networks~\cite{gast2018lightweight}, predict confidence maps or aleatoric uncertainty to identify unreliable regions. However, most systems treat uncertainty as a passive output for post-hoc analysis rather than as an active component of the inference process. Recent studies have revisited Bayesian filtering principles in neural inference to regulate iterative dynamics~\cite{liu2024revisiting}. 
While variational Bayesian techniques (e.g., VBReg~\cite{Jiang_2023_CVPR}) have been explored for 3D point cloud outlier rejection, their application in dense 2D spatial alignment is hindered by the highly non-convex nature of image matching. Our work aims to bridge this gap, evolving from traditional 3D outlier rejection to a fully recurrent 2D feature-manifold estimation under large appearance gaps.
\section{Methodology}
\label{sec:method}

\begin{figure*}[t]    
\centering    
\includegraphics[width=\textwidth]{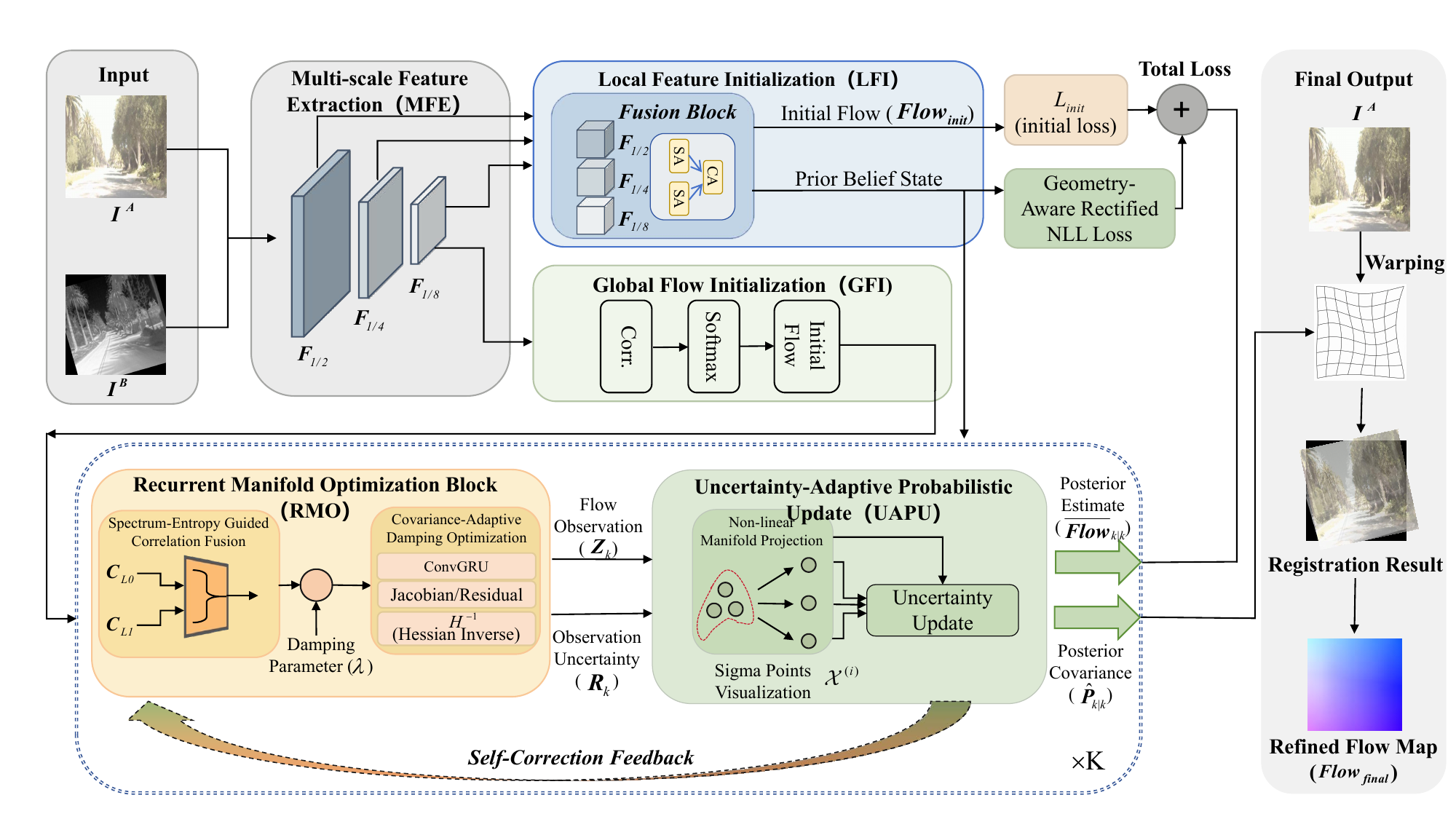}     
\caption{\textbf{Overview of the proposed RBE-Flow architecture. } 
RBE-Flow formulates the alignment task as a closed-loop recurrent Bayesian estimation on feature manifolds. 
\textbf{(a) MFE. }A weight-shared CNN encoder extracts features at 1/2, 1/4, 1/8 scale from the input image pair.
\textbf{(b) LFI. }A combination of SA and CA mechanisms fuses the features from different scales.
\textbf{(c) GFI. }In this module, we use 4D correlation volume on 1/8 scale feature to generate the initial flow field.
\textbf{(d) RMO. }This module iteratively generates flow residuals by solving a damped feature-metric least-squares problem, where the damping factor is adaptively regularized by the recirculated posterior uncertainty.
\textbf{(e) UAPU. }This module propagates deterministic Sigma points through the manifold to compute an MMSE-optimal update. It assimilates optimization-derived likelihoods into the prior state, while updating the posterior covariance to guide subsequent RMO.
}
\label{fig:overview}
\end{figure*}
RBE-Flow constructs a closed-loop recurrent Bayesian estimation framework, aiming to calculate feature flows in complex cross-modal scenarios. We formulate flow estimation as a dynamic state estimation problem rather than a direct regression. 
The overall pipeline begins with a multi-scale feature encoder to provide hierarchical semantic representations for subsequent optimization. The core recurrent process consists of two coupled modules: RMO, which generates the flow observation by solving a local non-linear optimization problem on the feature manifold; UAPU, which assimilates the observation with the prior belief to compute the posterior distribution. The posterior uncertainty output by UAPU is fed back to RMO to modulate the trajectory of the next optimization, establishing a robust self-correction mechanism.
An overview of our framework is shown in Fig.~\ref{fig:overview}.

\subsection{Multi-scale Feature Extraction (MFE)}
To better capture the texture features of cross-modal image pairs, we utilize a weight-shared ResNet-based CNN encoder to extract multi-scale features at 1/2, 1/4, and 1/8 scales. 
The 1/8 scale features capture global semantic context and remain modality-independent. Therefore, we pass the 1/8 scale features into the Iteration Initialization Module to generate a robust initial flow. The features at 1/2 and 1/4 scales retain rich spatial details and texture information, which effectively compensate for the detail loss in deeper features.

These multi-scale features constitute the foundation of the system: coarse-scale features are used for the initialization stage, while fine-scale features serve as local observation evidence in the subsequent recurrent Bayesian inference.

\subsection{Iteration Initialization Module}
Before entering the recurrent state estimation loop, constructing a reliable Prior Belief State is paramount. In this module, we aim to initialize both the feature representations and the flow field.

\noindent\textbf{Local Feature Initialization (LFI).}
To generate features containing both high-level semantics and rich low-level textures, we design a multi-scale feature fusion strategy. 
We fuse the 1/8 scale and 1/4 scale features via self-attention (SA) to inject global context into mid-level representations. Subsequently, cross-image Cross-Attention (CA) is applied to enhance cross-modal correspondences. This process is repeated with the 1/2 scale features, yielding a distilled feature manifold that preserves high-resolution structural evidence for the likelihood computation.

\noindent\textbf{Global Flow Initialization (GFI).}
Although ground-truth geometric displacements in cross-modal datasets are often synthesized via global affine matrices, we deliberately adopt a dense, pixel-wise flow formulation. Directly estimating low-dimensional global parameters makes the regression highly sensitive to cross-modal noise; a minuscule variance in parameter space inevitably triggers a drastic amplification of geometric errors across peripheral regions. By modeling pixel-wise dense correspondences on the manifold instead, our framework robustly absorbs local radiometric anomalies and mitigates error propagation. So to provide a robust initial estimation for subsequent iterative refinement, we compute the initial flow utilizing the 1/8 scale features $\boldsymbol{F}_0$ and $\boldsymbol{F}_1$.
We estimate the dense correlation matrix $\mathbf{C} \in \mathbb{R}^{HW \times HW}$ between feature maps $\boldsymbol{F}_0$ and $\boldsymbol{F}_1$:
\begin{equation}
    \boldsymbol{C}(\mathbf{u}, \mathbf{v}) = \frac{\boldsymbol{F}_0(\mathbf{u})^\top \boldsymbol{F}_1(\mathbf{v})}{\sqrt{d}},
\end{equation}
where $\mathbf{u}, \mathbf{v}$ denote pixel coordinates and $d$ is the channel dimension.

To extract dense correspondences, we normalize the correlation into a matching probability distribution $\mathbf{P}$ via softmax, and then compute the expected coordinate $\hat{\mathbf{v}}$ on the target image:
\begin{equation}
\boldsymbol{P}(\mathbf{u}, \mathbf{v}) = \frac{\exp(\boldsymbol{C}(\mathbf{u},\mathbf{v}))}{\sum_{\mathbf{v}' } \exp(\boldsymbol{C}(\mathbf{u},\mathbf{v}'))},
\end{equation}
\begin{equation}
    \quad \hat{\mathbf{v}}(\mathbf{u}) = \sum_{\mathbf{v}} \boldsymbol{P}(\mathbf{u}, \mathbf{v}) \cdot \mathbf{v}.
\end{equation}

The initial flow $\boldsymbol{Flow}_{init}$ is derived as:
\begin{equation}
    \boldsymbol{Flow}_{\text{init}}(\mathbf{u}) = \hat{\mathbf{v}}(\mathbf{u}) - \mathbf{u}.
\end{equation}

This initial optical flow is passed into RMO, constituting the initial prior mean for the recurrent Bayesian estimation process and marking the starting point of the closed-loop estimation.

\subsection{Recurrent Manifold Optimization Block (RMO).}
Based on the initialized flow, the system enters the recurrent loop. 
In the $k$-th iteration, this module serves as the observation generator of our Bayesian estimation. It solves a local non-linear optimization on the manifold to yield a flow observation $\boldsymbol{Z}_k$, and its associated uncertainty $\mathbf{R}_k$.
The data pair $(\boldsymbol{Z}_k, \boldsymbol{R}_k)$ constitutes the complete parameters of the likelihood distribution and serves as input for probabilistic fusion in the subsequent UAPU module.

Our method is based on a core assumption: high-dimensional image features are not randomly scattered throughout the ambient space, but rather reside on a lower-dimensional non-linear manifold. Therefore, flow estimation is essentially finding the shortest path on this feature manifold that connects the corresponding features between the image pair.

\noindent\textbf{Spectrum-Entropy Guided Correlation Fusion. }
To capture the multi-scale geometric structure of the manifold, we construct a two-level correlation volume: $\mathbf{C}_{L0}$(local window) for high-frequency details; and $\boldsymbol{C}_{L1}$(pooled features) for low-frequency global consistency. To suppress redundant noise, we perform feature fusion via Spectrum Entropy-Guided Fusion (SGCF). First, we calculate the entropy $\mathcal{H}$ for $\boldsymbol{C}_{L0}$ and $\boldsymbol{C}_{L1}$ respectively from the correlation probability distribution $\mathbf{p}$:
\begin{equation}
    \mathcal{H}(\mathbf{u}) = -\sum_{i} p_i \log p_i, 
\end{equation}
where $\mathbf{p} = \operatorname{softmax}(\boldsymbol{C}_{\text{L}}).$

Subsequently, adaptive fusion weights are computed through a lightweight SpecNet $\Phi_{\text{Spec}}$:
\begin{equation}
    [\omega_{L0}, \omega_{L1}] = \operatorname{softmax}(\Phi_{\text{Spec}}([\boldsymbol{C}_{L0}, \boldsymbol{C}_{L1}, \mathcal{H}_{L0}, \mathcal{H}_{L1}])).
\end{equation}

The final fused correlation :
\begin{equation}
    \boldsymbol{C}_{\text{fused}} = \omega_{L0} \cdot \boldsymbol{C}_{L0} + \omega_{L1} \cdot \boldsymbol{C}_{L1}.
\label{eq:spectral_fusion}
\end{equation}
This mechanism enables the network to dynamically adjust its reliance on different scale information.

\noindent\textbf{Covariance-Adaptive Damping Optimization (CDO). }
The core task of RMO is to find $\Delta \boldsymbol{Flow}$ to minimize the locally linearized least-squares objective function. We utilize a multi-head ConvGRU to directly predict the Jacobian $\boldsymbol{J}_k$, residual gradient $\mathbf{g}_k$, and damping measurement $\lambda_k$ required for the subsequent LM computation. 
For $\lambda$, we introduce the posterior covariance from the previous timestep to adjust the damping coefficient:
\begin{equation}
    \lambda'_k = \lambda_k + \beta \cdot \operatorname{tr}(\boldsymbol{P}_{k-1|k-1}).
\end{equation}
When the prior uncertainty is high, $\lambda'_k$ increases, adaptively switching the update from a Gauss-Newton step toward a more conservative Gradient Descent.
The flow increment $\Delta \boldsymbol{Flow}_k$ is solved via:
\begin{equation}
    \left( \boldsymbol{J}_k^\top \boldsymbol{J}_k + \lambda'_k \boldsymbol{I}\right) \Delta \boldsymbol{Flow}_k = -\boldsymbol{J}_k^\top \mathbf{g}_k.
\label{eq:lm_update}
\end{equation}

We define the observation result:
\begin{equation}
    \boldsymbol{Z}_k = \hat{\boldsymbol{Flow}}_{k-1} + \Delta \boldsymbol{Flow}_k.
\label{eq:observation_model}
\end{equation}

\subsection{Uncertainty-Adaptive Probabilistic Update (UAPU)}
The observations of RMO are accurate on a local cost surface, but may exhibit errors (such as occlusions and motion boundaries) from a global perspective. 
The core idea of UAPU is to model the flow refinement as a Non-linear Belief Propagation problem: at each iteration step, the system takes the posterior distribution of the previous timestep as the prior belief, fuses it with the likelihood evidence provided by RMO, and computes the current posterior distribution via Bayes' theorem, thereby achieving self-correction of errors.

\noindent\textbf{Uncertainty-Diffused Prior Belief Propagation.}
We model the flow state transition as a random walk. The prior mean $\hat{\boldsymbol{Flow}}_{k|k-1}$ is:
\begin{equation}
    \hat{\boldsymbol{Flow}}_{k|k-1} = \hat{\boldsymbol{Flow}}_{k-1|k-1}.
\end{equation}

Considering that flow uncertainty exhibits significant spatial heterogeneity across different regions, we introduce QNet $\Psi_Q$ to adaptively predict the process noise $\boldsymbol{Q}_k$, yielding the prior covariance $\boldsymbol{P}_{k|k-1}$:
\begin{equation}
    \quad \boldsymbol{P}_{k|k-1} = \boldsymbol{P}_{k-1|k-1} + \boldsymbol{Q}_k.
\label{eq:prior_cov}
\end{equation}

To ensure numerical stability and positive semi-definiteness, we apply factor $\mathbf{S}$ as the Cholesky decomposition result of $\boldsymbol{P}$.

\noindent\textbf{Non-linear Manifold Distribution Projection.}
Due to the strong non-linearity of the feature manifold, performing local linearization on the mapping from state to observation space would introduce non-negligible truncation errors.
We generate $2L+1 = 5$ deterministic sampling points $\boldsymbol{X}^{(i)}$ around the prior mean $\hat{\boldsymbol{Flow}}_{k|k-1}$: 1 center point, and $2L$ perturbed points symmetrically distributed along the axes of the state manifold. These are projected into the observation space to obtain the predicted observation point set $\mathbf{Z}_k^{(i)}$:
\begin{equation}
    \mathbf{Z}^{(i)}_k =\operatorname{ObsNet}( \boldsymbol{X}^{(i)}_k),
\end{equation}
where $i \geq 1$ denotes the $i$-th Sigma point. The observation network $\operatorname{ObsNet}(\cdot)$ acts as a learnable non-linear mapping function to bridge the state space and the feature-metric observation space, which is implemented as a lightweight MLP network composed of $1 \times 1$ convolutions.

The predicted observation mean is calculated as:
\begin{equation}
    \hat{\boldsymbol{Z}}_k = \sum_{i=0}^{4} W_m^{(i)} \mathbf{Z}_k^{(i)},
\end{equation}
where $W_m^{(0)} = \frac{\lambda_{\text{scale}}}{L + \lambda_{\text{scale}}}$,$L = 2$ characterizes the spatial dimensionality of the optical flow vector, $\lambda_{\text{scale}}$ acts as a scaling parameter governing the manifold spread of the generated Sigma points.

\noindent\textbf{Uncertainty-Gated Posterior Belief Fusion.}
To optimally fuse the prior belief with the observation evidence, we compute the innovation covariance $\boldsymbol{P}_{zz}$ and the state-measurement cross-covariance $\boldsymbol{P}_{xz}$ via the weighted outer product of the sampling points:
\begin{equation}
    \boldsymbol{P}_{zz}= \sum_{i=0}^{4} W_c^{(i)} (\boldsymbol{Z}^{(i)}_k - \hat{\boldsymbol{Z}}_k)(\boldsymbol{Z}^{(i)}_k - \hat{\boldsymbol{Z}}_k)^\top + \boldsymbol{R}_k,
\end{equation}
where $\boldsymbol{R}_k$ is the observation noise covariance estimated from a light ConvNet.
\begin{equation}
    \boldsymbol{P}_{xz}= \sum_{i=0}^{4} W_c^{(i)} (\boldsymbol{X}^{(i)}_k - \hat{\boldsymbol{Flow}}_{k|k-1})(\boldsymbol{Z}^{(i)}_k - \hat{\boldsymbol{Z}}_k)^\top.
\label{eq:covariances}
\end{equation}

Under the Gaussian approximation assumption of the posterior distribution, minimizing the mean square error $\operatorname{tr}(\mathbf{P}_{k|k})$ of the posterior estimation is equivalent to solving the following unconstrained optimization for $\mathcal{K}_k$. Thus, under the MMSE criterion, we obtain the optimal Bayesian fusion gain $\mathcal{K}_k$:
\begin{equation}
\frac{\partial\, \operatorname{tr}(\boldsymbol{P}_{k|k})}{\partial \mathcal{K}_k} = \mathbf{0} \implies \mathcal{K}_k = \boldsymbol{P}_{xz}\boldsymbol{P}_{zz}^{-1}.
\label{eq:kalman_gain}
\end{equation}

The posterior estimate is updated following Bayes' theorem, fusing the prior and likelihood:
\begin{equation}
    \hat{\boldsymbol{Flow}}_{k|k} = \hat{\boldsymbol{Flow}}_{k|k-1} + \mathcal{K}_k (\boldsymbol{Z}_k - \hat{\boldsymbol{Z}}_k),
\end{equation}
and the posterior covariance are updated as:
\begin{equation}
    \boldsymbol{P}_{k|k} = \boldsymbol{P}_{k|k-1} - \mathcal{K}_k \boldsymbol{P}_{zz} \mathcal{K}_k^\top.
\end{equation}
This update restricts the information gain by contracting the prior uncertainty space. 

Finally, the posterior statistical moments $(\hat{\boldsymbol{Flow}}_{k|k}, \boldsymbol{S}_{k|k})$ output by UAPU are recirculated, forming a complete closed-loop recurrent system. This feedback mechanism establishes a Bayesian inference chain in the temporal dimension, enabling the system to progressively converge the uncertain initial belief to an accurate posterior approximation of the true feature flow through iterative evidence accumulation.

\subsection{Loss Function}
To effectively train this recurrent probabilistic framework, we adopt a multi-stage loss, bridging the deterministic global alignment and the stochastic local refinement. We construct an Initialization Loss to guide the network for preliminary alignment at coarse resolution, and a Geometry-Aware Rectified NLL Loss.

\noindent\textbf{Initialization Loss.}
For the initial flow $\boldsymbol{Flow}_{\text{init}}$ generated in GFI, we employ a simple L1 loss to supervise it. We downsample the optical flow ground truth $\boldsymbol{Flow}_{\text{gt}}$ to the same resolution as $\boldsymbol{Flow}_{\text{init}}$ and compute the L1 distance between them:
\begin{equation}
\mathcal{L}_{\text{init}} = \frac{1}{N} \sum_{\mathbf{p}} \left| \boldsymbol{Flow}_{\text{init}} - \boldsymbol{Flow}_{\text{gt}} \right|,
\end{equation}
where $N$ is the number of pixels. This loss ensures that the model has a reasonable global alignment foundation before entering the fine matching stage.

\noindent\textbf{Geometry-Aware Rectified NLL Loss.}
The vanilla NLL objective frequently exhibits numerical instability, often leading to degenerated variance estimates; models exploit gradients by predicting practically zero variances ($\sigma \to 0$) to minimize the penalty, leading to exploding gradients.
To enforce a meaningful uncertainty estimate, we introduce a Geometry-Aware Rectification.
We impose a dynamic lower bound on the predicted variance, scaling with the magnitude of the geometric error:
\begin{equation}
    (\hat{\sigma}^{(j)}_k)^2 = \operatorname{softplus}(s^{(j)}_k) + \alpha \cdot ( \hat{\boldsymbol{Flow}}^{(j)}_k - \boldsymbol{Flow}_{\text{gt}}^{(j)} )^2 + \epsilon,
\label{eq:rectified_variance}
\end{equation}
here, $\mathbf{S}$ is the Cholesky factor output by UAPU, i.e., the square root of the covariance matrix. 
This mechanism ensures that high-error regions are forced to have high uncertainty, preventing the "overconfident but wrong" pathology.
The Negative Log-Likelihood loss in iteration $k$ is:
\begin{equation}
    \mathcal{L}_{\text{NLL}}^{(k)} = \frac{1}{2} \sum_{j \in \{x,y\}} \left( \log (\hat{\sigma}^{(j)}_k)^2 + \frac{(\hat{\boldsymbol{Flow}}^{(j)}_k - \boldsymbol{Flow}_{\text{gt}}^{(j)})^2}{(\hat{\sigma}^{(j)}_k)^2} \right).
    \label{eq:nll_loss}
\end{equation}

Finally, we apply deep supervision on the outputs of the $N_{\text{iter}}$ iterations, and compute their weighted sum using a decay coefficient $\gamma=0.9$, realizing a loss function that is both result-oriented and process-oriented.
\begin{equation}
    \mathcal{L}_{\text{NLL}} = \frac{1}{\sum_{k=1}^{N_{\text{iter}}} \gamma^{N_{\text{iter}}-k}} \sum_{k=1}^{N_{\text{iter}}} \gamma^{N_{\text{iter}}-k} \mathcal{L}_{\text{NLL}}^{(k)}, 
    \label{eq:fine_loss_sum}
\end{equation}

\noindent\textbf{Total Loss.}
Finally, we perform a weighted sum of the Initialization Loss and the Geometry-Aware Rectified NLL Loss to obtain the total loss function:
\begin{equation}
    \mathcal{L}_{\text{total}} = \lambda_{\text{init}} \mathcal{L}_{\text{init}} + \lambda_{\text{NLL}} \mathcal{L}_{\text{NLL}}.
\end{equation}
where $\lambda_{\text{init}}$ and $\lambda_{\text{NLL}}$ are the respective weights for the two losses.
\section{Experiments}
\label{sec:experiments}
\subsection{Implementation Details}
RBE-Flow is initialized using the publicly released XoFTR pretrained weights (trained at 640 resolution), and then its entire framework is fine-tuned end-to-end.
We train this framework on RoadScene~\cite{xu2023murf,xu2020aaai}, WHU-OPT-SAR~\cite{ji2018fully}, and OSdataset~\cite{xiang2020automatic}, with a learning rate of $3\times {10^{-4}}$ and a batch size of 8.

\noindent\textbf{Datasets.}
We conduct experiments on OSdataset, an Optical-SAR dataset covering a wide variety of terrain types such as forests, farmlands, and rivers; WHU-OPT-SAR dataset, an Optical-SAR dataset focused on complex urban scenes; and RoadScene dataset, a RGB-IR dataset capturing street views and traffic scenarios under varying lighting conditions. RoadScene is derived from the MURF ~\cite{xu2022murf} through rigorous cropping and expansion to ensure robust correspondence learning.
We follow the standard splits of 1172/153/288 (train/val/test) for OSdataset, 7000/700/700 for WHU-OPT-SAR dataset, 5120/640/640 for RoadScene dataset.
Our data augmentation strategy introduces random scaling between $[0.9,1.1]$, rotations within $[-30^\circ,30^\circ]$, and translations of up to 15 pixels. During the training process, images are resized to $64 \times 64$ patches. Furthermore, to maintain consistent supervision across all stages, the ground-truth flow fields are directly derived from the provided pixel-wise alignments.

\noindent\textbf{Evaluation Metrics.}
We quantify the matching efficacy using the Average End-Point Error (AEPE) and the Correct Match Rate (CMR). While AEPE computes the expected flow displacement divergence, CMR indicates the percentage of samples that successfully meet the particular error constraint. 
Guided by existing evaluation protocols, CMR is computed over an interval of thresholds (between 0.1 and 5 pixels) to yield a multi-angle geometric precision under different constraints.

\noindent\textbf{Baselines.}
Inspired by MatchAnything~\cite{he2025matchanything} and CoMatch~\cite{zhang2025comatcher}, RBE-Flow is compared with such representative methods from the following categories:
\begin{itemize}    
    \item \textit{Sparse matching methods: }     
    LNIFT~\cite{li2022lnift},     
    RIFT2~\cite{li2019rift, li2023rift2},
    HOWP~\cite{zhang2023histogram}, 
    MSG~\cite{zheng2025msg}. 
    Such handcrafted keypoint-descriptor methods provide stable alignments under structural noise. But their vulnerability to cross-modal appearance gaps and large-scale geometric shifts severely limits their applicability.
    
    \item \textit{Semi-dense matching methods:}      
    XoFTR~\cite{tuzcuouglu2024xoftr}.      
    Serving as a prime example of contemporary Transformer-driven, detector-free architectures, its matching capability noticeably deteriorates when faced with severe textural discrepancies across modalities.
    
    \item \textit{Dense matching methods:}             
    RAFT~\cite{teed2020raft}, 
    GMFlow~\cite{xu2023unifying},  
    ADRNet~\cite{xiao2024adrnet},   
    GDROS~\cite{sun2025gdros}.         
    Such methods generate dense pixel-level flows. However, general flow estimators struggle profoundly with massive modality gaps, whereas adapted models (e.g., GDROS and ADRNet) remain restricted by domain-specific priors and low-resolution outputs. 
\end{itemize}

\subsection{Experiment Results}

\begin{figure*}[t]    
\centering    
\includegraphics[width=\textwidth]{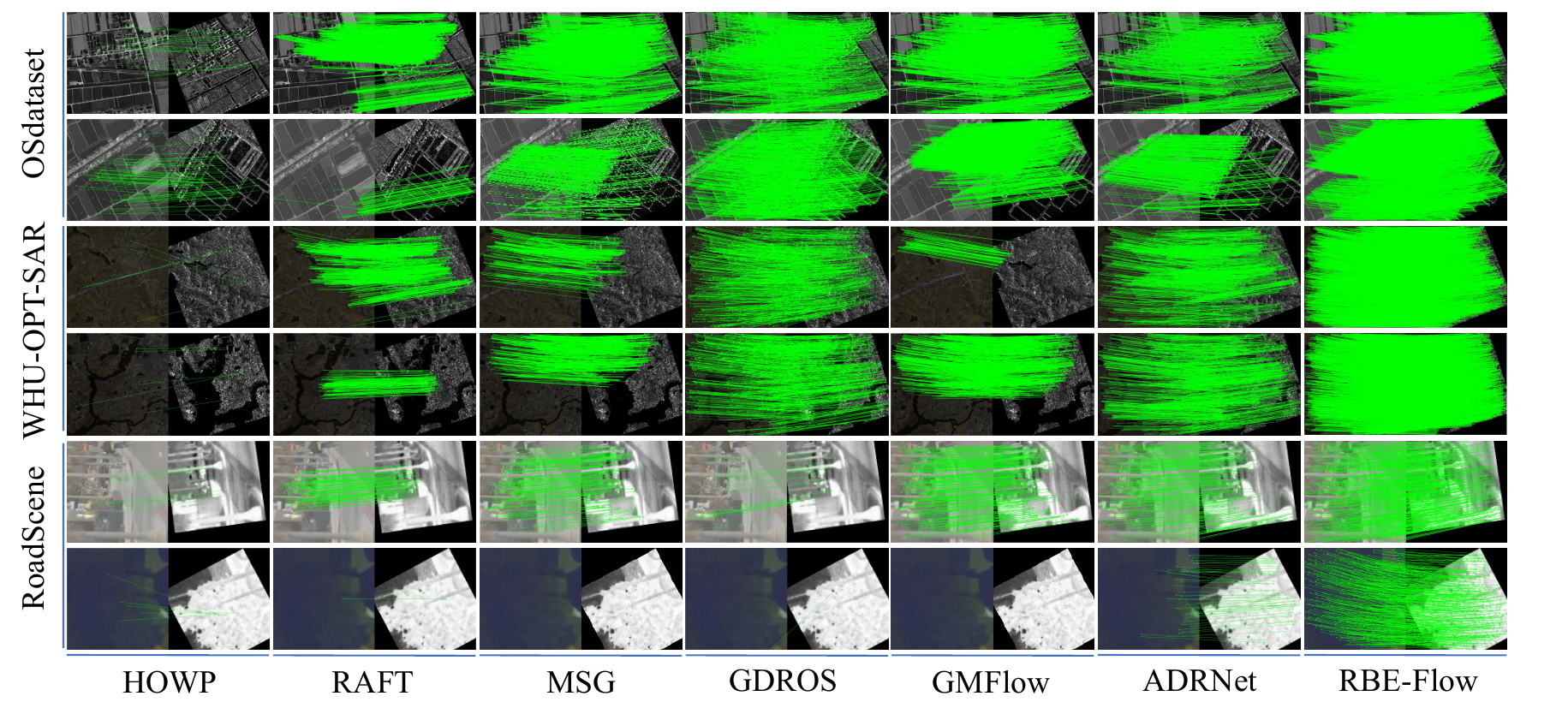}     
\caption{\textbf{Inlier correspondence visualization with state-of-the-art methods. }For each image pair, we uniformly sample 5000 candidate correspondences, and only the matches with registration error less than 2 pixels are visualized. As illustrated by the inlier visualizations, RBE-Flow consistently achieves the most comprehensive and geometrically stable correspondences across all tested benchmarks. The dense matching results underscores the efficacy of our recurrent Bayesian framework in resolving profound modality gaps and structural distortions.}    
\label{fig:pointfig}
\end{figure*}
We compare RBE-Flow with 9 representative handcrafted, sparse, semi-dense, and dense correspondence methods and fine-tuned learning-based methods on the three datasets respectively. All of the experiments show our RBE-Flow achieves the best result even under strict thresholds. This demonstrates the effectiveness of our method in cross-modal registration tasks.

\begin{figure*}[t]
    \centering
    \begin{subfigure}[b]{0.49\textwidth}
        \centering
        \includegraphics[width=\linewidth]{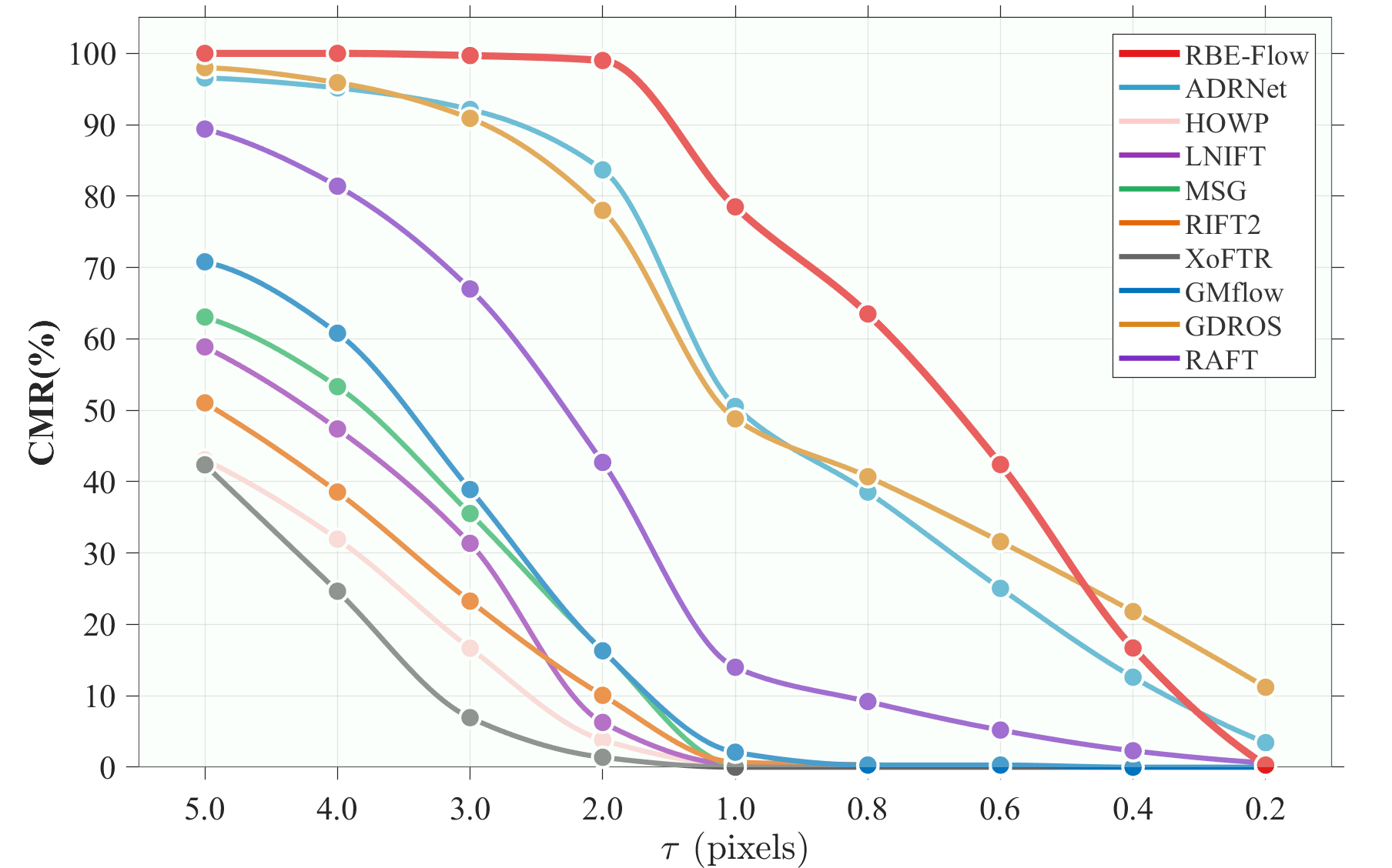}
        \caption{OSdataset}
        \label{fig:line_os}
    \end{subfigure}
    \begin{subfigure}[b]{0.49\textwidth}
        \centering
        \includegraphics[width=\linewidth]{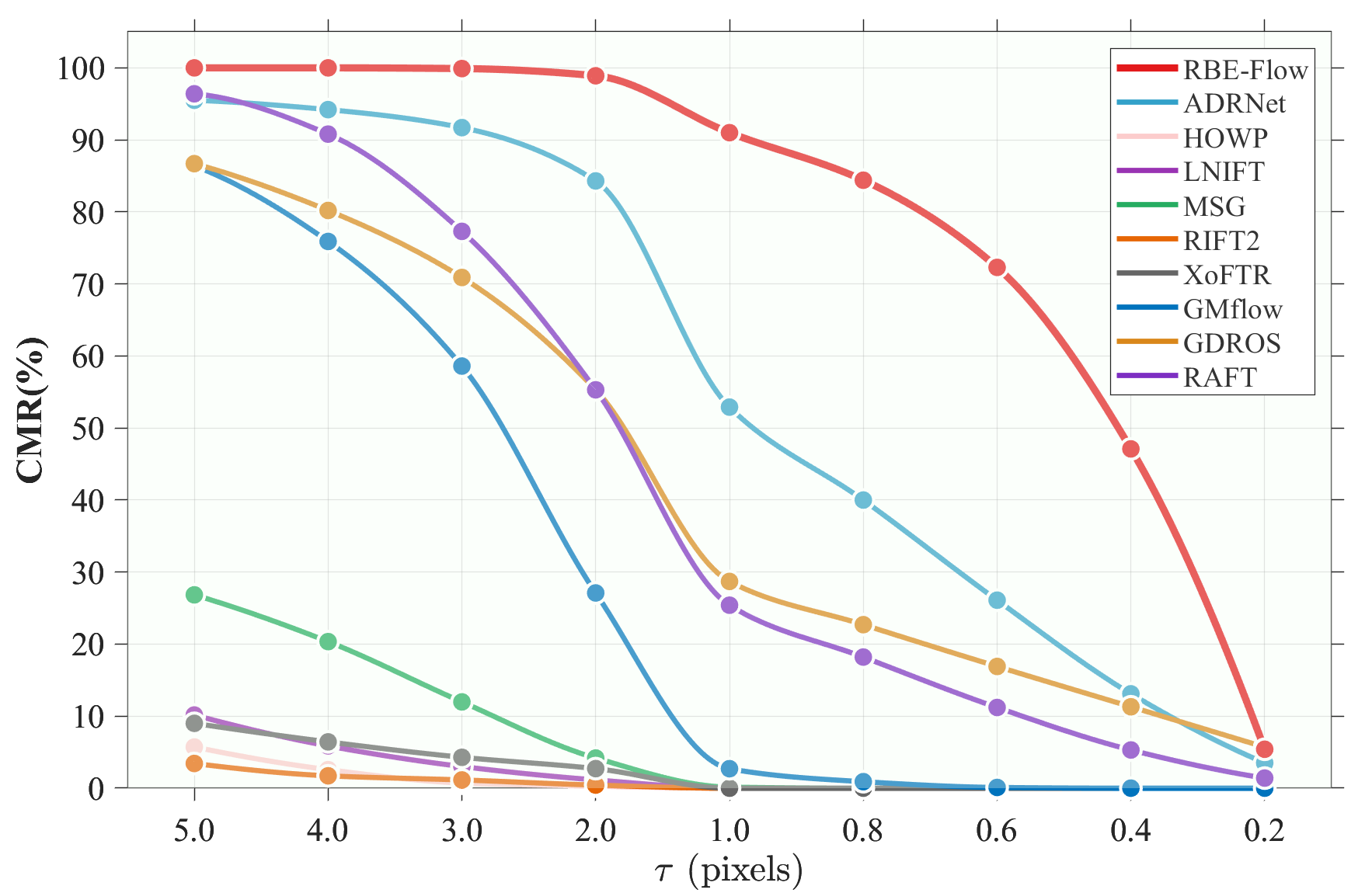} 
        \caption{WHU-OPT-SAR dataset}
        \label{fig:line_whu}
    \end{subfigure}
    \begin{subfigure}[b]{0.49\textwidth}
        \centering
        \includegraphics[width=\linewidth]{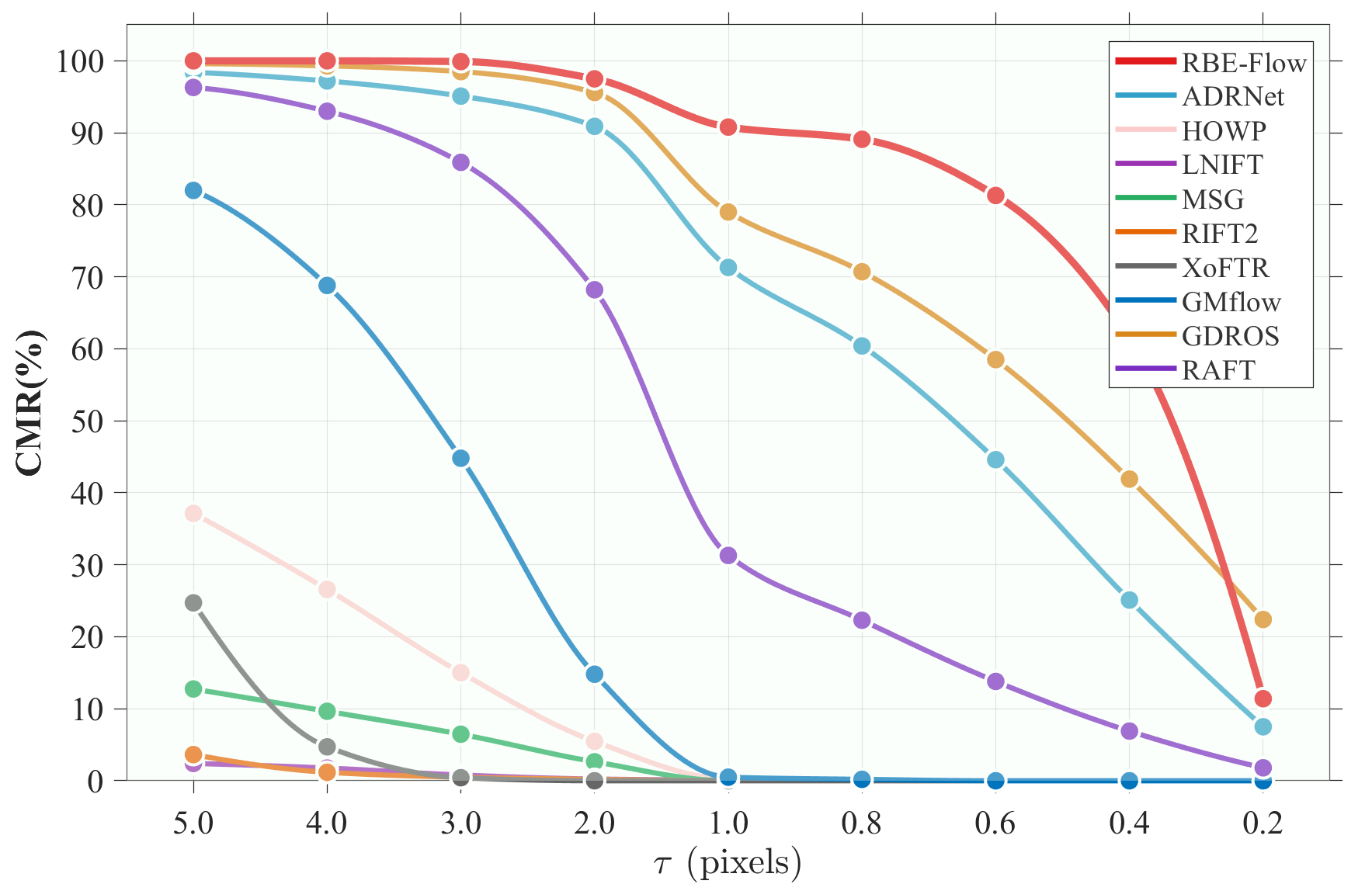} 
        \caption{RoadScene dataset}
        \label{fig:line_rs}
    \end{subfigure}
    
    \caption{\textbf{CMR curves under varying thresholds on (a) OSdataset, (b) WHU-OPT-SAR, and (c) RoadScene. }We compare RBE-Flow with representative handcrafted, sparse, and (semi-)dense matching methods by evaluating the CMR across a range of pixel thresholds. RBE-Flow consistently achieves the highest CMR under both strict and relaxed thresholds, demonstrating its robustness to cross-modal appearance discrepancies and geometric variations.}
    \label{fig:all_line_charts}
\end{figure*}

\noindent\textbf{Results on OSdataset.}
As shown in Table~\ref{tab:all_datasets_results} and Fig.~\ref{fig:all_line_charts}(a), our RBE-Flow demonstrates highly competitive performance on the OSdataset. Compared to the previous state-of-the-art method ADRNet, our approach significantly reduces the AEPE from $1.41px$ to $0.77px$, achieving a remarkable $45.4\%$ error reduction. In terms of CMR, RBE-Flow has the highest accuracy at both normal and fine thresholds, achieving $78.5\%$ and $57.3\%$ at $1px$ and $0.7px$, respectively. Although our method ranks second at the most stringent threshold of $0.3px$ ($11.6\%$) trailing slightly behind GDROS ($16.6\%$), the qualitative line chart in Fig.~\ref{fig:all_line_charts}(a) indicates that RBE-Flow consistently maintains a superior overall trend across a continuous range of error tolerances.

\noindent\textbf{Results on WHU-OPT-SAR.}
In this dataset, our method shows the most prominent accuracy. RBE-Flow outperforms all methods in all thresholds. Specifically, it achieves an AEPE of 0.53, nearly a $3\times$ improvement over the second-best ADRNet ($1.43$). For the CMR metric, while traditional methods (e.g., HOWP, RIFT2) and some deep learning approaches completely fail on this challenging dataset, our RBE-Flow achieves extreme robustness with a CMR of $91.0\%$ at $1px$ and $80.0\%$ at $0.7px$ (outperforming ADRNet's $52.9\%$ and $33.0\%$). Fig.~\ref{fig:all_line_charts}(b) visually confirms this dominance, where the RBE-Flow retains a high matching ratio even when the threshold becomes extraordinarily strict.

\noindent\textbf{Results on RoadScene.}
RBE-Flow achieves the best overall performance on the RoadScene dataset. As presented in Table~\ref{tab:all_datasets_results}, it shows the state-of-the-art AEPE of 0.49px. The steep and rapid decay of the baseline curves in Fig.~\ref{fig:all_line_charts}(c) further underscores the difficulty of this dataset, yet RBE-Flow remains notably stable against varying error constraints.

\noindent\textbf{Overall Discussion.}
The series of experiments show the two main advantages of RBE-Flow: strong cross-domain generalization and high sub-pixel accuracy. While recent baselines such as ADRNet and RAFT perform reasonably well on the OSdataset, they both struggle with the WHU-OPT-SAR and RoadScene dataset. Moreover, as shown in the line charts (Fig.~\ref{fig:all_line_charts}), existing methods share a common limitation: their performance drops sharply when the error threshold is strictly reduced ($t < 1px$). RBE-Flow largely mitigates this issue, maintaining a robust matching ratio even under tight tolerances. This stability implies that our method provides more reliable fine-grained alignments across varying real-world conditions.
\begin{table*}[htbp]
\centering
\small 
\caption{\textbf{Quantitative comparison on OSdataset, WHU-OPT-SAR, and RoadScene datasets. }
AEPE and CMR at thresholds $t=\{3,1,0.7,0.3\}$. 
Best results are in \textbf{bold}, and second-best are \underline{underlined}.}
\label{tab:all_datasets_results}
\resizebox{\textwidth}{!}{%
\begin{tabular}{l ccccc ccccc ccccc}
\toprule
\multicolumn{1}{c}{\multirow{3}{*}{Methods}} & 
\multicolumn{5}{c}{OSdataset} & 
\multicolumn{5}{c}{WHU-OPT-SAR} & 
\multicolumn{5}{c}{RoadScene} \\
\cmidrule(lr){2-6} \cmidrule(lr){7-11} \cmidrule(lr){12-16}
& \multicolumn{1}{c}{\multirow{2}{*}{AEPE $\downarrow$}} & \multicolumn{4}{c}{CMR $\uparrow$ (\%)} 
& \multicolumn{1}{c}{\multirow{2}{*}{AEPE $\downarrow$}} & \multicolumn{4}{c}{CMR $\uparrow$ (\%)} 
& \multicolumn{1}{c}{\multirow{2}{*}{AEPE $\downarrow$}} & \multicolumn{4}{c}{CMR $\uparrow$ (\%)} \\
\cmidrule(lr){3-6} \cmidrule(lr){8-11} \cmidrule(lr){13-16}
& & $@3px$ & $@1px$ & $@0.7px$ & $@0.3px$ & & $@3px$ & $@1px$ & $@0.7px$ & $@0.3px$ & & $@3px$ & $@1px$ & $@0.7px$ & $@0.3px$ \\
\midrule
HOWP~\cite{zhang2023histogram}~{\scriptsize ISPRS'23} & 20.39 & 16.7 & 0.3 & 0.0 & 0.0 & 157.14 & 0.7 & 0.0 & 0.0 & 0.0 & 16.67 & 15.0 & 0.2 & 0.1 & 0.0 \\
LNIFT~\cite{li2022lnift}~{\scriptsize TGRS'22} & 46.98 & 31.3 & 0.3 & 0.0 & 0.0 & 167.00 & 3.0 & 0.0 & 0.0 & 0.0 & 28.89 & 0.8 & 0.1 & 0.0 & 0.0 \\
MSG~\cite{zheng2025msg}~{\scriptsize TGRS'25} & 44.50 & 35.5 & 0.3 & 0.0 & 0.0 & 120.97 & 12.0 & 0.1 & 0.0 & 0.0 & 43.59 & 6.5 & 0.0 & 0.0 & 0.0 \\
RIFT2~\cite{li2019rift}~{\scriptsize TIP'20} & 29.83 & 23.3 & 0.7 & 0.3 & 0.0 & 214.08 & 1.1 & 0.0 & 0.0 & 0.0 & 29.45 & 0.5 & 0.0 & 0.0 & 0.0 \\
GMFlow~\cite{xu2023unifying}~{\scriptsize TPAMI'23} & 5.13 & 38.9 & 2.1 & 0.3 & 0.0 & 3.57 & 27.1 & 2.7 & 0.6 & 0.0 & 4.42 & 44.8 & 0.5 & 0.1 & 0.0 \\
XoFTR~\cite{tuzcuouglu2024xoftr}~{\scriptsize CVPR'24} & 5.41 & 6.9 & 0.0 & 0.0 & 0.0 & 18.05 & 4.3 & 0.0 & 0.0 & 0.0 & 6.00 & 0.4 & 0.0 & 0.0 & 0.0 \\
RAFT~\cite{teed2020raft}~{\scriptsize ECCV'20} & 2.89 & 67.0 & 14.0 & 7.0 & 1.3 & 2.07 & 77.3 & 25.4 & 14.6 & 3.0 & 1.79 & 85.9 & 31.3 & 17.9 & 4.0 \\
ADRNet~\cite{xiao2024adrnet}~{\scriptsize TGRS'24} & \underline{1.41} & \underline{92.5} & \underline{50.5} & 31.9 & 7.5 & \underline{1.43} & \underline{91.7} & \underline{52.9} & \underline{33.0} & 7.7 & 0.98 & 95.1 & 71.3 & 53.1 & 15.6 \\
GDROS~\cite{sun2025gdros}~{\scriptsize TGRS'25} & 1.51 & 78.0 & 48.8 & \underline{36.2} & \textbf{16.6} & 2.87 & 70.9 & 28.7 & 19.8 & \underline{8.5} & \underline{0.68} & \underline{98.5} & \underline{79.0} & \underline{65.1} & \underline{32.4} \\
\midrule
\textbf{RBE-Flow (ours)} & \textbf{0.77} & \textbf{99.7} & \textbf{78.5} & \textbf{57.3} & \underline{11.6} & \textbf{0.53} & \textbf{99.9} &  \textbf{91.0} & \textbf{80.0} & \textbf{24.0} & \textbf{0.49} & \textbf{99.9} &\textbf{90.8} & \textbf{85.8} & \textbf{37.7} \\
\bottomrule
\end{tabular}%
}
\end{table*}

\subsection{Ablation Study}
To validate the effectiveness of our proposed recurrent Bayesian estimation framework, we ablate the core components on the same dataset with Sec.4.2. We define our baseline (A) as the multi-scale CNN encoder appended with LFI, and calculated following an effective, coarse-to-fine, semi-dense point matching. Then we gradually add GFI, our loss function (loss), RMO, and UAPU. Results are shown in Table~\ref{tab:ablation_3datasets_t103}.

\begin{table}[htbp]
\centering
\small 
\caption{\textbf{Ablation experiments on three datasets.}
AEPE and CMR at thresholds $t=\{1,0.7,0.3\}$.
Best results are in \textbf{bold}, and second-best are \underline{underlined}.}
\label{tab:ablation_3datasets_t103}

\resizebox{\textwidth}{!}{
\begin{tabular}{lccccccccccccc}
\toprule
\multicolumn{1}{c}{\multirow{3}{*}{Method}} &
\multicolumn{4}{c}{OSdataset} &
\multicolumn{4}{c}{WHU-OPT-SAR} &
\multicolumn{4}{c}{RoadScene} \\
\cmidrule(lr){2-5} \cmidrule(lr){6-9} \cmidrule(lr){10-13}
& \multicolumn{1}{c}{\multirow{2}{*}{AEPE $\downarrow$}} & \multicolumn{3}{c}{CMR $\uparrow$ (\%)} 
& \multicolumn{1}{c}{\multirow{2}{*}{AEPE $\downarrow$}} & \multicolumn{3}{c}{CMR $\uparrow$ (\%)} 
& \multicolumn{1}{c}{\multirow{2}{*}{AEPE $\downarrow$}} & \multicolumn{3}{c}{CMR $\uparrow$ (\%)} \\
\cmidrule(lr){3-5} \cmidrule(lr){7-9} \cmidrule(lr){11-13}
& & $@1px$ & $@0.7px$ & $@0.3px$ &  & $@1px$ & $@0.7px$ & $@0.3px$ &  & $@1px$ & $@0.7px$ & $@0.3px$ \\
\midrule
(1) A+L1
& 1.32 & 22.9 & 2.8 & 0.0
& 5.52 & 2.5 & 1.2 & 0.0
& 0.69 & 87.4 & 76.0 & 1.0 \\
(2) A+L1+GFI
& 0.98 & 61.8 & 32.6 & 0.3
& 1.10 & 47.3 & 13.9 & 0.3
& 0.64 & 87.3 & 67.0 & 13.0 \\
(3) A+GFI+loss
& 0.91 & 66.7 & 36.8 & 0.7
& 0.76 & 82.9 & 58.6 & 1.0
& 0.59 & 89.0 & 78.9 & 12.6 \\
(4) A+GFI+loss+RMO
& \underline{0.84} & \underline{71.9} & \underline{49.0} & \underline{3.1}
& \underline{0.61} & \underline{89.3} & \underline{75.4} & \underline{11.9}
& \underline{0.54} & \underline{89.8} & \underline{81.4} & \underline{26.9} \\
(5) A+GFI+loss+RMO+UAPU 
& \textbf{0.77} & \textbf{78.5} & \textbf{57.3} & \textbf{11.6}
& \textbf{0.53} & \textbf{91.0} & \textbf{80.0} & \textbf{24.0}
& \textbf{0.49} & \textbf{90.8} & \textbf{85.8} & \textbf{37.7} \\
\bottomrule
\end{tabular}
}
\end{table}

\textbf{(1) Baseline (A+L1).}
The baseline performs poorly when the CMR within 1px, on the WHU-OPT-SAR dataset, it results in an AEPE of 5.52px and low accuracy at strict sub-pixel thresholds (0.7px is $1.2\%$), showing its limitation to estimate reliable cross-modal fine-grained correspondences.

\textbf{(2) A+L1+GFI.}
Adding the GFI module provides a reliable initial flow by computing dense feature correlations, which helps resolve large spatial displacements.
As a result, the performance improves significantly. The AEPE on the WHU-OPT-SAR dataset drops from 5.52 to 1.10, and the CMR($1px$) on the OSdataset increases from $22.9\%$ to $61.8\%$. This shows that a global correlation step is necessary to provide a good starting point for subsequent refinement.

\textbf{(3) A+GFI+loss.}
In this step, we replace the standard L1 refinement loss with our Geometry-Aware Rectified NLL Loss. The metrics reflect this improvement: the CMR under 1px on the WHU-OPT-SAR dataset increases from $47.3\%$ to $82.9\%$, demonstrating that properly constrained probabilistic supervision leads to better alignment.

\textbf{(4) A+GFI+loss+RMO.}
Introducing the RMO block allows the network to iteratively solve a local non-linear optimization problem on the feature manifold, computing flow increments dynamically rather than guessing them through a black-box layer.
This module particularly benefits fine-grained matching. On the RoadScene dataset, the strict sub-pixel accuracy (CMR:0.3px) improves from $12.6\%$ to $26.9\%$, showing the effectiveness of the covariance-adaptive damping mechanism in local detail refinement.

\textbf{(5) A+GFI+loss+RMO+UAPU.}
Finally, integrating the UAPU block forms our complete RBE-Flow framework. This closed-loop uncertainty feedback yields the best performance across all datasets. The improvement is especially notable at the strictest $0.3px$ threshold, increasing from $3.1\%$ to $11.6\%$ and from $11.9\%$ to $24.0\%$ on the OSdataset and WHU-OPT-SAR respectively.

\noindent \textbf{Summary.}
The ablations validate a clear coarse-to-fine strategy: while a robust global flow initialization from GFI is essential for resolving large displacements, it is the synergistic RMO-UAPU loop that enables superior sub-pixel precision by empowering the system to reason about its own posterior uncertainty, leading to the best overall performance across all datasets.

\noindent \textbf{Efficiency Analysis.}
RBE-Flow is trained end-to-end using a single NVIDIA RTX 4090 GPU with 14.3M parameters. During inference, RBE-Flow achieves an average runtime of 1.48 ms per image pair. Each iteration adds a cost of 0.13 ms. Specifically, during the training phase (batch size of 8, 7 recurrent iterations), our full probabilistic framework (RMO+UAPU) adds a modest temporal overhead of only 29.42 ms per step ($+12.9\%$) compared to the deterministic RMO-only baseline. Crucially, the memory overhead introduced by the UAPU block is virtually negligible ($+0.002$ GB), keeping the total peak GPU memory strictly bounded at 10.16 GB. This thoroughly validates that our recurrent Bayesian paradigm achieves high-precision multi-modal alignment without incurring prohibitive computational or memory overhead.

\section{Conclusion}
\label{sec:conclusion}
We present RBE-Flow, a closed-loop recurrent Bayesian estimation framework specifically designed for robust cross-modal registration. By reformulating flow estimation as a dynamic state estimation problem rather than a standard regression, our approach explicitly addresses the inherent ambiguities in cross-modal feature matching.
The architecture is fundamentally driven by the tight coupling of localized manifold optimization, which acts as a source of flow observations, and a recurrent Bayesian update mechanism, which sequentially refines the state distribution through non-linear belief propagation.
Experimental results on the OSdataset, WHU-OPT-SAR, and RoadScene datasets demonstrate that RBE-Flow consistently achieves state-of-the-art performance particularly under strict sub-pixel thresholds.
Looking forward, the principles of recurrent Bayesian estimation and uncertainty-driven feedback introduced in RBE-Flow offer a path toward more reliable spatial alignment in high-stakes applications like long-term autonomous navigation and cross-modality medical diagnosis. The flexibility of our modular design ensures that RBE-Flow remains compatible with evolving large-scale vision models and advanced geometric reasoning techniques, facilitating more robust cross-modal perception in increasingly complex environments. In future work, we plan to extend this probabilistic manifold optimization strategy from 2D dense matching to 3D multi-modal scene reconstruction.
\\
\\
\noindent\textbf{Acknowledgment.} This work was supported by Hebei Natural Science Foundation: F2025501003, 
and the Fundamental Research Funds for the Central Universities: N2523014.
Shijiazhuang-Northeastern University Science and Technology Cooperation Special Project: NEUS2025-01-003.


\bibliographystyle{splncs04}
\bibliography{main}
\end{document}